\documentclass[11pt]{article}

\PassOptionsToPackage{table}{xcolor}
\usepackage[preprint]{acl}

\usepackage{times}
\usepackage{latexsym}
\usepackage[T1]{fontenc}
\usepackage[utf8]{inputenc}
\usepackage{microtype}
\usepackage{inconsolata}
\usepackage{graphicx}
\usepackage{adjustbox}
\usepackage{subcaption}
\usepackage{caption}
\usepackage{placeins} 
\usepackage{float}   

\usepackage{amsmath,amssymb,amsthm}
\usepackage{booktabs}
\usepackage{multirow}
\usepackage{algorithm}
\usepackage{algorithmic}

\newcommand{\conv}[1]{#1\textsuperscript{\textdagger}}
\newcommand{\postconv}[1]{\textcolor{black!55}{#1}}

\title{Active Learners as Efficient PRP Rerankers}
\author{
  \normalfont
  Jerem\'ias Figueiredo Paschmann,
  Juan Kaplan,
  Francisco Nattero, \\
  Santiago Mauricio Barron Bucolo,
  Juan Wisznia,
  Luciano del Corro \\[0.5em]
  \footnotesize
  \texttt{\{jfigueiredopaschmann,jkaplan,fnattero,sbarronbucolo,jwisznia,delcorrol\}@udesa.edu.ar} \\[0.5em]
  ELIAS Lab, Departamento de Ingenier\'ia, Universidad de San Andr\'es
}

\begin{document}
\maketitle

\begin{abstract}
Pairwise Ranking Prompting (PRP) elicits pairwise preference judgments from an LLM, which are then aggregated into a ranking, usually via classical sorting algorithms.
However, judgments are noisy, order-sensitive, and sometimes intransitive, so sorting assumptions don't match the setting. Because sorting aims to recover a full permutation, truncating it to meet a call budget does not produce a dependable top-$K$. We thus reframe PRP reranking as active learning from noisy pairwise comparisons and show active rankers are drop-in replacements that improve NDCG@10 per call in the call-constrained regime.
Our noise-robust framework also introduces a randomized-direction oracle that uses a single LLM call per pair. This approach converts systematic position bias into zero-mean noise, enabling unbiased aggregate ranking without the cost of bidirectional calls.\footnote{Code available at \url{https://github.com/jerecoder/IReranker}}
\end{abstract}

\section{Introduction}
LLMs are increasingly used for reranking in Retrieval-Augmented Generation (RAG): given a query and a candidate list, rerankers aggregate LLM pairwise preferences into an ordered top-$K$ subset that strongly affects downstream answer quality
\citep{zhou2025large, zhu2023llmir,dong2024grag,sun2025dynamicrag}.
Major cloud providers now offer reranking as managed services, making call efficiency a first-class concern: LLM invocations dominate cost and latency, and the goal is a sorted prefix, not the full list.\looseness=-1

Commonly, PRP is paired with classical sorting algorithms
\citep{qin2024pairwise,sun2023chatgptsearch}: PRP supplies noisy preference judgments while sorting determines which pairs to query.
This is structurally mismatched: sorting assumes transitive comparisons, while LLM judgments are stochastic and can violate transitivity.
Sorting thus wastes budget polishing an unstable permutation rather than improving the top-$K$.\looseness=-1

LLM order effects, where swapping document presentation order can flip the judge's choice between documents, further complicate matters \citep{shi2024judges,yin2025fragile,jeong2025comparativetrap}.
Standard PRP queries both prompt directions at 2 calls per pair \citep{qin2024pairwise,wu2025realtimeprp}, yet preference cycles persist.

We therefore frame PRP reranking as active learning from noisy pairwise comparisons, choosing adaptively which pairs to query to maximize top-$K$ quality within a budget. This connects to the literature on best-$K$ identification under stochastic feedback
\citep{mohajer2017active,heckel2016activeranking,shah2018simplerobust,ren2020bestk,luo-etal-2024-prp}.
We also evaluate a cheaper oracle: randomizing the prompt direction yields a one-call estimate that converts position bias into zero-mean noise.\looseness=-1

We study two questions:
\textbf{(Q1)} \emph{Does active ranking outperform state-of-the-art PRP rerankers at fixed budget (NDCG@10)?}
\textbf{(Q2)} \emph{Does randomized-direction prompting improve the NDCG@10--cost trade-off beyond scheduling alone?}
The best-performing active scheduler in our experiments is the algorithm of \citet{mohajer2017active}, which we call \textbf{Mohajer}: it adaptively selects which pairs to query, concentrating comparisons near the top-$K$ boundary.
\textbf{Q1}: On TREC DL2019/2020 with Flan-T5-XL, Mohajer outperforms the best sorting baseline by $+$9.7 NDCG@10 at $B{=}300$ calls (66.1 vs.\ 56.4), under the same bidirectional oracle, with the advantage holding across the entire call-constrained regime ($B{=}200$--$450$).
\textbf{Q2}: Randomized-direction prompting improves both strategies, but in different ways.
For PRP rerankers, it raises quality at fixed budget: BubbleSort gains $+$5.5 NDCG@10 at $B{=}300$ (56.4$\to$62.0) simply by halving the call cost per pair and covering more comparisons.
For active rankers, the effect is more pronounced: comparing Mohajer under both oracles, the randomized-direction oracle raises the quality ceiling from 66.96 to 68.0 while reducing the calls needed to reach it from $B{=}450$ to $B{=}250$, a 44\% reduction.
Across BEIR-style tasks, active rankers reach NDCG@10 comparable to QuickSort (Avg.\ 56.8 for Flan-T5-XL) with up to 7$\times$ fewer calls.

\section{Related Work}

\noindent\textbf{Pairwise LLM reranking.}
PRP elicits pairwise preferences from an LLM and aggregates them into a ranking \citep{sun2023chatgptsearch,qin2024pairwise}, typically via sorting algorithms that assume transitivity and target an unbudgeted complete order.

\noindent\textbf{Order effects.}
LLM comparisons are direction-sensitive \citep{shi2024judges,yin2025fragile,jeong2025comparativetrap}, so PRP often queries both prompt orders, doubling cost \citep{qin2024pairwise,wu2025realtimeprp}.
Our randomized-direction oracle makes one call per pair, producing aggregate outcomes robust to direction bias.

\noindent\textbf{PRP beyond sorting.}
PRP-Graph uses adaptive pairings \citep{luo-etal-2024-prp} and tournament designs structure comparisons \citep{chen2024tourrank}.
We reframe reranking as active learning from noisy feedback and evaluate active top-$K$ identifiers as drop-in replacements for sorting \citep{heckel2016activeranking,shah2018simplerobust,ren2020bestk,mohajer2017active}, using adaptive pairings in the spirit of PRP-Graph but grounded in noise-tolerant active ranking theory.

\noindent\textbf{Complementary paradigms.}
Setwise and listwise methods~\citep{zhuang2024setwise,huang2025tssetrank,wang2025realm} reduce cost by processing multiple documents per call, changing the prompting primitive itself; pairwise and listwise calls differ in token cost, context length, and bias, making raw call counts incommensurable across paradigms.
Our goal is to improve scheduling \emph{within} pairwise PRP, which remains widely deployed for its fine-grained signal and constrained-output reliability~\citep{qin2024pairwise}; the two directions are complementary.

\begin{table*}[t]
\centering
\begin{adjustbox}{max width=\textwidth}
\scriptsize
\begin{tabular}{c l rrrrrrrrr}
\toprule
& & \multicolumn{9}{c}{Number of LLM calls (budget)} \\
\cmidrule(lr){3-11}
Oracle & Ranker & 100 & 150 & 200 & 250 & 300 & 350 & 400 & 450 & 500 \\
\midrule

\multirow{6}{*}{\rotatebox[origin=c]{90}{Bidirectional}}
& BubbleSort       & \underline{49.27} & \underline{49.27} & \underline{56.43} & \underline{56.43} & 56.42 & 56.98 & 60.25 & 60.30 & 60.51 \\
& HeapSort          & 6.81 & 6.13 & 6.13 & 9.28 & 23.04 & 41.84 & 54.29 & 62.81 & \textbf{68.21} \\
& QuickSort         & \textbf{55.93} & \textbf{55.89} & 55.87 & 56.20 & 56.20 & 56.20 & 56.40 & 56.59 & 56.68 \\
& PAC + Bubble      & \underline{49.27} & \underline{49.27} & 49.27 & 49.27 & \underline{57.52} & \underline{60.59} &
\conv{60.61} & \postconv{60.61} & \postconv{60.61} \\
& Mohajer + Bubble  & 30.12 & 30.12 & \textbf{62.34} & \textbf{64.80} & \textbf{66.09} & \textbf{66.28} & \textbf{66.83} &
\conv{\textbf{67.02}} & \postconv{\underline{67.02}} \\
& Mohajer           & 30.12 & 30.12 & \textbf{62.34} & \textbf{64.80} & \textbf{66.09} & \textbf{66.28} & \underline{66.81} &
\conv{\underline{66.96}} & \postconv{66.96} \\

\midrule

\multirow{6}{*}{\rotatebox[origin=c]{90}{Randomized}}
& BubbleSort        & \underline{55.90$\pm$0.28} & 56.10$\pm$0.21 & 59.82$\pm$0.20 & 59.68$\pm$0.21 & 61.95$\pm$0.21 & 62.03$\pm$0.18 & 64.04$\pm$0.19 & 64.00$\pm$0.25 & 65.42$\pm$0.30 \\
& HeapSort          & 6.58$\pm$0.14 & 16.46$\pm$0.43 & 50.17$\pm$0.24 & 65.80$\pm$0.16 & \textbf{68.50$\pm$0.39} & \textbf{68.41$\pm$0.30} & \textbf{68.34$\pm$0.11} & \textbf{68.53$\pm$0.21} & \textbf{68.71$\pm$0.21} \\
& QuickSort         & 54.49$\pm$0.27 & 54.71$\pm$0.23 & 55.26$\pm$0.24 & 56.20$\pm$0.47 & 57.56$\pm$0.19 & 58.95$\pm$0.34 & 59.81$\pm$0.41 & 61.87$\pm$0.36 & 63.76$\pm$0.38 \\
& PAC + Bubble      & 49.27$\pm$0.00 & \underline{57.02$\pm$0.21} &
\conv{\underline{60.01$\pm$0.21}} & \postconv{60.01$\pm$0.21} & \postconv{60.01$\pm$0.20} & \postconv{60.01$\pm$0.21} & \postconv{60.01$\pm$0.20} & \postconv{60.01$\pm$0.21} & \postconv{60.01$\pm$0.21} \\
& Mohajer + Bubble  & \textbf{61.36$\pm$0.31} & \textbf{65.84$\pm$0.33} & \textbf{67.66$\pm$0.35} & \underline{67.59$\pm$0.34} & \underline{68.14$\pm$0.26} & \underline{68.25$\pm$0.27} &
\conv{\underline{68.08$\pm$0.12}} & \postconv{\underline{68.08$\pm$0.12}} & \postconv{\underline{68.08$\pm$0.12}} \\
& Mohajer           & \textbf{61.36$\pm$0.31} & \textbf{65.84$\pm$0.33} &
\textbf{67.66$\pm$0.35} & \conv{\textbf{68.00$\pm$0.19}} & \postconv{68.00$\pm$0.19} &
\postconv{68.00$\pm$0.19} & \postconv{68.00$\pm$0.19} & \postconv{68.00$\pm$0.19} & \postconv{68.00$\pm$0.19} \\

\bottomrule
\end{tabular}%
\end{adjustbox}
\caption{Average NDCG@10 (\%) on TREC DL 2019 and DL 2020 with Flan-T5-XL across LLM call budgets. Bold = best per column; underline = second-best (within each oracle block). \textsuperscript{\textdagger} indicates the smallest budget at which a method completes; results at larger budgets are de-emphasized. \textbf{Randomized} rows report mean $\pm$ 95\% bootstrap CI half-width over 8 oracle seeds (10k resamples); \textbf{Bidirectional} is deterministic given outcomes so CIs are omitted.}
\label{tab:budget_ndcg10_oracles_flan_xl}
\end{table*}

\section{Reranking from Noisy Comparisons}
\label{sec:setup}

Given a query $q$, a first-stage retriever returns $N$ candidates
$\mathcal{D}(q)=\{d_1,\dots,d_N\}$ ($N\ge K$). The reranker outputs an ordered
top-$K$ list $\mathcal{R}_K(q)=(r_1,\dots,r_K)$ with $r_\ell \in \mathcal{D}(q)$.

\noindent\textbf{Pairwise oracle interface.}
Algorithms interact with candidates only via noisy pairwise outcomes: for each unordered pair $\{i,j\}$, a call returns $X_{ij}(q)\in\{0,1\}$, where $X_{ij}(q)=1$ means $d_i$ is preferred (judged more relevant to $q$) over $d_j$, i.e.\ $d_i \succ d_j$, with win probability $p_{ij}(q):=\Pr[X_{ij}(q)=1]$.
We assume only \emph{pair-consistency}, $p_{ij}(q)=1-p_{ji}(q)$ for $i\neq j$ (this is enforced via oracle design).

\noindent\textbf{Call-centric cost.}
We count LLM inference calls: bidirectional uses two per pair, randomized-direction uses one. Since calls dominate PRP cost \citep{wisznia2025batchingcaching}, this changes which routines are optimal.

\vspace{-0.2cm}
\paragraph{Oracles} Let $\text{LLM}(d_a, d_b) \in \{1, 0\}$ denote the outcome of one call, where $1$ means the first document is preferred.

\noindent\textbf{Bidirectional (two calls).}
The standard PRP oracle \citep{qin2024pairwise}: $V_{ij}=1$ iff $\text{LLM}(d_i,d_j)=1 \wedge \text{LLM}(d_j,d_i)=0$, else $V_{ij}=0$.

\noindent\textbf{Randomized-direction (one call).}
We randomize input order: $V_{ij}=\text{LLM}(d_i,d_j)$ with probability $1/2$, else $V_{ij}=1-\text{LLM}(d_j,d_i)$. This ensures reciprocity in expectation, i.e.\ $\Pr[V_{ij}{=}1] = 1 - \Pr[V_{ji}{=}1]$: each individual call may be position-biased, but averaging over the random direction converts systematic bias into zero-mean noise, preserving pair-consistency (proof in Appendix~\ref{sec:unbiased-proof}).

\section{Selecting Active Rankers for Call-Budgeted Top-$K$ Reranking}
\label{sec:select_active_rankers}

Sorting treats every comparison as equally informative. Under a budget, this uniformity is wasteful. Active rankers concentrate comparisons on candidates whose relative order remains uncertain. This is the key mechanism behind our gains: a better schedule for the same comparator, requiring only lightweight bookkeeping with no model training or forward passes. The dominant cost remains the LLM call itself.

Our goal is high-quality top-$K$ prefixes under a strict call budget $B$ via the pairwise-oracle interface of \S\ref{sec:setup}.
We select algorithms with three criteria:
\textbf{(C1)} \emph{Top-$K$ objective}: targets best-$K$/prefix identification;
\textbf{(C2)} \emph{Noise tolerance}: well-defined under pair-consistency without assuming a global order;
\textbf{(C3)} \emph{Anytime behavior}: outputs a competitive top-$K$ prefix as comparisons accrue.

We focus on comparison scheduling gains and benchmark two complementary active rankers: tournament-based vs.\ anchor-based. Methods assuming transitivity or targeting a full global ranking are omitted.

\noindent\textbf{Tournament/heap extraction.}
\citet{mohajer2017active} identifies best-$K$ via tournaments with heap extraction, focusing comparisons on likely contenders (C1--C3). We use one oracle call per match.

\noindent\textbf{Anchor-based Probably Approximately Correct (PAC) best-$K$.}
\citet{agarwal2022pac} identifies best-$K$ via anchors and winner sets (C1, C3).
We take anchors from a zero-cost BM25 prior and restrict comparisons to the top $K{\times}m$ ($m{=}3$) BM25 prefix, keeping calls low.

\noindent\textbf{Ordered outputs.}
PAC returns an unordered best-$K$ set, so we apply BubbleSort on the final top-$K$. Mohajer outputs an ordered prefix; BubbleSort polishing is optional and negligible. Any added comparisons count toward the budget.
\begin{table*}[t]
\centering
\scriptsize
\setlength{\tabcolsep}{2.0pt}        
\renewcommand{\arraystretch}{1.00}   
\resizebox{0.85\textwidth}{!}{
\begin{tabular}{llccccccccc}
\toprule
& \textbf{Ranker} & \textbf{Covid} & \textbf{Robust04} & \textbf{Touche} & \textbf{SciFact} & \textbf{DBPedia} & \textbf{DL19} & \textbf{DL20} & \textbf{Avg. NDCG@10} & \textbf{Avg. Calls/Task}\\
\midrule
& \textbf{BM25} & 59.5 & 40.7 & 44.2 & 67.9 & 31.9 & 50.6 & 48.0 & 49.0 & - \\
\midrule

\multirow{9}{*}{\rotatebox{90}{\textbf{Flan-T5-L}}}
& BubbleSort@10 (Bidirectional) & 70.9 & \textbf{44.2} & \textbf{44.7} & \textbf{69.2} & \textbf{41.7} & 63.4 & 58.6 & \textbf{56.1} & 679 \\
& HeapSort (Bidirectional)   & 76.0 & 40.4 & 33.2 & 67.5 & \underline{41.4} & \textbf{65.0} & \textbf{62.6} & \underline{55.2} & 1230 \\
& QuickSort (Bidirectional)  & 76.2 & 41.0 & 27.4 & 60.1 & 41.1 & \underline{64.5} & 58.5 & 52.7 & 1954 \\
\arrayrulecolor{black!35}
\cmidrule(lr){2-11}
\arrayrulecolor{black}
& PAC+Bubble (Bidirectional) & 69.3 & \underline{44.0} & \underline{41.4} & \underline{68.5} & 39.2 & 61.7 & 57.2 & 54.5 & 323 \\
& PAC+Bubble (Randomized)    & 70.2 & 41.0 & 38.2 & 67.0 & 38.1 & 60.0 & 57.3 & 53.1 & 184 \\
& Mohajer+Bubble (Bidirectional) & \underline{76.5} & 37.8 & 26.7 & 54.9 & 40.0 & 63.0 & 56.4 & 50.7 & 423 \\
& Mohajer+Bubble (Randomized)    & \textbf{76.9} & 37.8 & 25.7 & 58.8 & 40.0 & 61.8 & \underline{58.8} & 51.4 & 354 \\
& Mohajer (Bidirectional)        & 76.5 & 37.5 & 26.4 & 53.8 & 39.7 & 62.6 & 56.1 & 50.4 & 399 \\
& Mohajer (Randomized)           & 76.2 & 36.2 & 24.4 & 57.5 & 39.1 & 60.6 & 57.2 & 50.2 & 232 \\
\midrule

\multirow{9}{*}{\rotatebox{90}{\textbf{Flan-T5-XL}}}
& BubbleSort@10 (Bidirectional) & 74.8 & \textbf{55.4} & \textbf{42.8} & \textbf{71.3} & \textbf{43.1} & 68.4 & 67.0 & \textbf{60.4} & 941 \\
& HeapSort (Bidirectional)   & \underline{78.2} & \underline{54.9} & 28.4 & \underline{70.6} & \underline{41.6} & \textbf{70.6} & \textbf{68.9} & \underline{59.0} & 1409 \\
& QuickSort (Bidirectional)  & 77.2 & 53.7 & 25.8 & 61.4 & \underline{41.6} & \underline{70.4} & 67.2 & 56.8 & 1669 \\
\arrayrulecolor{black!35}
\cmidrule(lr){2-11}
\arrayrulecolor{black}
& PAC+Bubble (Bidirectional) & 71.3 & 48.9 & \underline{41.1} & 70.4 & 39.1 & 62.6 & 58.6 & 56.0 & 332 \\
& PAC+Bubble (Randomized)    & 71.3 & 48.1 & 38.8 & 68.6 & 38.8 & 61.1 & 58.5 & 55.0 & 184 \\
& Mohajer+Bubble (Bidirectional) & 76.0 & 53.7 & 25.7 & 61.9 & 40.9 & 66.6 & 67.5 & 56.0 & 427 \\
& Mohajer+Bubble (Randomized)    & \textbf{78.5} & 54.0 & 27.9 & 63.5 & 41.2 & 69.5 & 66.3 & 57.3 & 345 \\
& Mohajer (Bidirectional)        & 76.0 & 53.6 & 25.4 & 61.2 & 40.7 & 66.6 & 67.3 & 55.8 & 399 \\
& Mohajer (Randomized)           & 77.6 & 53.2 & 27.2 & 62.8 & 40.4 & 68.7 & \underline{67.6} & 56.8 & 232 \\
\bottomrule
\end{tabular}%
}
\caption{End-to-end BEIR-style NDCG@10 (\%) and average pairwise LLM calls. Bold = best per column; underline = second-best per column (within each model block).}
\label{tab:beir_main}
\end{table*}

\section{Results}

\vspace{-0.2cm}
\paragraph{Setup.}
We rerank the top $N{=}100$ BM25 candidates into an ordered top-$K$ list ($K{=}10$) and report NDCG@10 on BEIR-style tasks (Table~\ref{tab:beir_main}) and TREC DL2019/2020, capping each method at $B\in\{100,150,\dots,500\}$ LLM calls. The pairwise oracle uses Flan-T5-L/XL under (i) bidirectional and (ii) randomized-direction prompting. BubbleSort uses caching \citep{wisznia2025batchingcaching}. Additional Qwen results and code are in the appendix/repository.

\vspace{-0.2cm}
\paragraph{Main findings.}
Table~\ref{tab:budget_ndcg10_oracles_flan_xl} reports NDCG@10 on TREC DL2019/2020 (Flan-T5-XL) vs.\ budget $B$ (CIs and bootstrap tests in Appendix~\ref{sec:significance}).
\emph{(i)} In the call-constrained regime ($B \approx 200$--$450$), Mohajer outperforms PRP rerankers under the same oracle.
\emph{(ii)} Randomized-direction compresses ``time-to-quality'': Mohajer reaches peak quality by $B{=}250$.
\emph{(iii)} At high budgets, sorting catches up as global refinement pays off.
PAC lags because its two-phase design splits budget across objectives; Mohajer's tournament concentrates comparisons on likely top candidates. PAC benefits when the BM25 prior is strong (e.g., Touché).\looseness=-1

\vspace{-0.2cm}
\paragraph{Bidirectional oracle}
\label{sec:ar-vs-sorting}

\noindent\textbf{Low budgets: sorting is preferable.}
At $B \in \{100,150\}$, QuickSort reaches $\approx 55.9$ NDCG@10 while Mohajer is in warm-up ($30.1$). Below the warm-up threshold ($\sim$100 calls for $N{=}100$, $K{=}10$) sorting is preferable; above it, active ranking dominates.

\noindent\textbf{Call-constrained regime: active reranking is better.}
Mohajer leads from $B{=}200$ to $B{=}450$: at $B{=}300$, 66.09 vs.\ 56.42 (+9.67); at $B{=}350$, 66.28 vs.\ 56.98 (+9.30); at $B{=}450$, Mohajer+Bubble 67.02 vs.\ HeapSort 62.81 (+4.21).
Paired bootstrap tests (10k query resamples, $p{<}0.05$; Table~\ref{tab:sig_vs_bubble} in the appendix) confirm these gains are significant: under the randomized oracle, Mohajer+Bubble significantly outperforms BubbleSort at every budget; under bidirectional, from $B{=}200$ onward.

\noindent\textbf{High budgets: sorting can catch up.} At $B{=}500$, HeapSort (68.21) narrowly exceeds Mohajer+Bubble (67.02) as global refinement pays off.

\vspace{-0.2cm}
\paragraph{Randomized oracle: faster ``time-to-quality'' for active ranking}
\label{sec:stochastic-oracle-vs-sorting}

Using one call per pair, randomized-direction covers $\sim$$2\times$ as many pairs as bidirectional at the same budget.
Mohajer is already strong at $B{=}100$ (61.4) and converges at 68.0 by $B{=}250$, suggesting that broader pair coverage may outweigh spending two calls per pair to reduce order effects.
HeapSort surpasses Mohajer at $B{=}300$ (68.50 vs.\ 68.00), reaching 68.71 at $B{=}500$; sorting remains preferable once budgets are large enough for global refinement to dominate.

\vspace{-0.2cm}
\paragraph{End-to-end efficiency in the full pipeline}
\label{sec:end_to_end}

Table~\ref{tab:beir_main} reports end-to-end results ($N{=}100$, $K{=}10$). For Flan-T5-XL, PRP baselines use 941--1669 calls/task (Avg.\ 56.8--60.4 NDCG@10), while Mohajer and PAC use 184--345 calls (Avg.\ 55.0--57.3), a $\sim$3--5$\times$ reduction.
Randomized-direction further cuts calls: Mohajer drops from 399 to 232 per task, making adaptive scheduling with randomized oracle a strong low-cost design.
Touché is an outlier where BM25 is strong and neural reranking headroom is limited.\looseness=-1

\vspace{-0.2cm}
\paragraph{Order effects and comparisons.}
Bidirectional prompting reverses the preferred document on 20.6\% of pairs, confirming substantial order effects.
Mohajer+Bubble achieves competitive or better NDCG@10 than PRP-Graph with fewer comparisons as model size increases.
A top-$k$ sweep shows the crossover where global refinement becomes preferable moves earlier as $k$ increases. Details in the appendix.

\vspace{-0.2cm}
\paragraph{Latency}
\label{sec:latency}

Latency is estimated as a sequential upper bound: inference time per comparison times total comparisons, ignoring parallelism (see Appendix~\ref{sec:parallelization}).
Figure~\ref{fig:limit_time_all_flan_a100} shows active ranking reaching strong quality earlier (Mohajer at 23.3s, PAC at 10.1s), with sorting overtaking only at long runtimes.
Both active rankers support within-query parallelism (independent tournaments / anchor comparisons), potentially reducing wall-clock time by an order of magnitude. Qwen and H100/H200 results in the appendix.
\begin{figure}[H]
    \centering
    \includegraphics[width=1\linewidth]{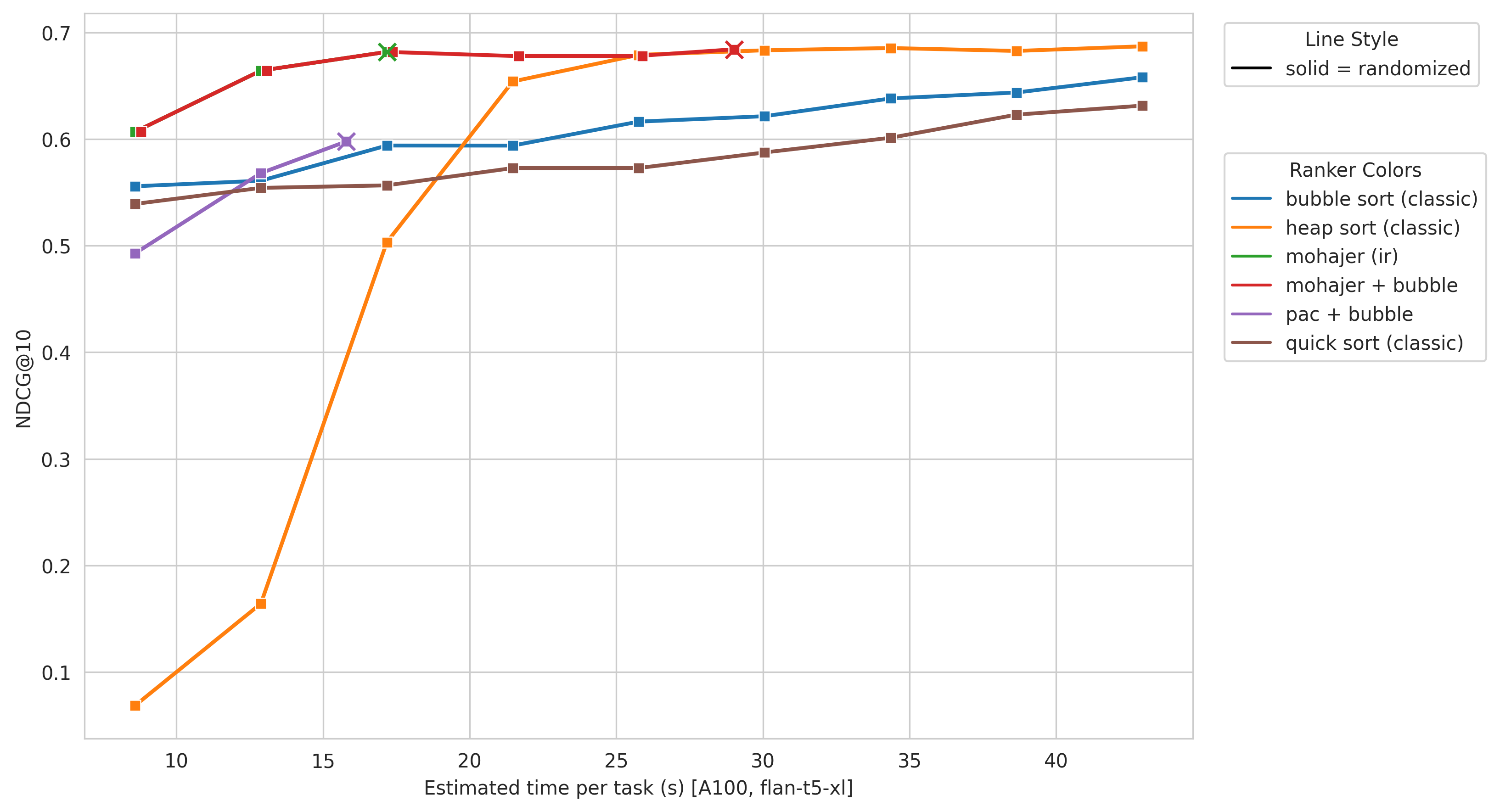}
    \caption{TREC DL 2019/2020 (Flan-T5-XL, A100): NDCG@10 vs.\ avg.\ time per task (randomized oracle). X marks the point at which a method has completed all its scheduled comparisons (convergence).}
    \label{fig:limit_time_all_flan_a100}
\end{figure}

\section{Conclusion}
We argue that PRP reranking is better modeled as budgeted learning from noisy pairwise comparisons than deterministic sorting.
Active rankers yield higher NDCG@10 at low budgets, while sorting helps mainly once large budgets make global refinement affordable.
Randomized-direction prompting further improves efficiency: at the same budget, it covers roughly twice as many pairs as bidirectional.
For practitioners deploying PRP in RAG pipelines, our results suggest a simple recipe: use Mohajer with the randomized-direction oracle when the call budget exceeds the warm-up threshold ($\sim$$K{\times}K$ calls), and fall back to sorting when budgets are either very small or large enough for global refinement.

\section*{Limitations}

Our study focuses on settings where a reliable pairwise LLM comparator can be elicited with constrained outputs. Results may vary with prompt design, model family, and decoding settings. Our cost metric counts LLM calls but omits system-level overheads (batching, network latency); latency measurements are not fully end-to-end. Parallel execution was not implemented, though both algorithms naturally support it (Appendix~\ref{sec:parallelization}).

The NDCG@10 gains from randomized-direction oracles are empirically consistent but not theoretically explained; the hypothesis that independent single-direction samples benefit adaptive algorithms more than correlated bidirectional samples is plausible but unproven.

Our PAC best-$K$ method (PAC+Bubble) introduces a candidate pool multiplier hyperparameter $m$ (default $m=3$), which controls the trade-off between comparison cost and coverage of the prior ranking. We did not perform a systematic ablation over $m$ due to computational constraints, and the optimal value likely depends on prior quality and dataset characteristics. While $m=3$ yields strong empirical results in our experiments, future work should investigate data-driven or adaptive selection strategies for this parameter.

Finally, active ranking theory often assumes conditional independence of oracle outputs; real LLM APIs can violate this through hidden state, caching, or nonstationarity.

\FloatBarrier
\bibliography{custom}

\begin{thebibliography}{21}
\providecommand{\natexlab}[1]{#1}

\bibitem[{Agarwal et~al.(2022)Agarwal, Khanna, and Patil}]{agarwal2022pac}
Arpit Agarwal, Sanjeev Khanna, and Prathamesh Patil. 2022.
\newblock \href {https://proceedings.mlr.press/v151/agarwal22a.html} {Pac
  top-$k$ identification under sst in limited rounds}.
\newblock In \emph{Proceedings of The 25th International Conference on
  Artificial Intelligence and Statistics}, volume 151 of \emph{Proceedings of
  Machine Learning Research}, pages 6814--6839. PMLR.

\bibitem[{Chen et~al.(2024)Chen, Liu, Zhang, Sun, Shi, Mao, and
  Yin}]{chen2024tourrank}
Yiqun Chen, Qi~Liu, Yi~Zhang, Weiwei Sun, Daiting Shi, Jiaxin Mao, and Dawei
  Yin. 2024.
\newblock \href {https://doi.org/10.48550/arXiv.2406.11678} {Tourrank:
  Utilizing large language models for documents ranking with a
  tournament-inspired strategy}.
\newblock \emph{CoRR}, abs/2406.11678.

\bibitem[{Dong et~al.(2024)Dong, Fatemi, Perozzi, Yang, and
  Tsitsulin}]{dong2024grag}
Jialin Dong, Bahare Fatemi, Bryan Perozzi, Lin~F. Yang, and Anton Tsitsulin.
  2024.
\newblock \href {https://doi.org/10.48550/arXiv.2405.18414} {Don't forget to
  connect! improving rag with graph-based reranking}.
\newblock \emph{arXiv preprint arXiv:2405.18414}.

\bibitem[{Heckel et~al.(2016)Heckel, Shah, Ramchandran, and
  Wainwright}]{heckel2016activeranking}
Reinhard Heckel, Nihar~B. Shah, Kannan Ramchandran, and Martin~J. Wainwright.
  2016.
\newblock \href {https://doi.org/10.48550/arXiv.1606.08842} {Active ranking
  from pairwise comparisons and when parametric assumptions don't help}.
\newblock \emph{arXiv preprint arXiv:1606.08842}.

\bibitem[{Huang et~al.(2025)Huang, Madala, Niu, Hockenmaier, and
  Zhang}]{huang2025tssetrank}
Jerry Huang, Siddarth Madala, Cheng Niu, Julia Hockenmaier, and Tong Zhang.
  2025.
\newblock \href {https://doi.org/10.48550/arXiv.2511.01208} {Contextual
  relevance and adaptive sampling for {LLM}-based document reranking}.
\newblock \emph{CoRR}, abs/2511.01208.

\bibitem[{Jeong et~al.(2025)Jeong, Park, Hong, Lee, and
  Choo}]{jeong2025comparativetrap}
Hawon Jeong, ChaeHun Park, Jimin Hong, Hojoon Lee, and Jaegul Choo. 2025.
\newblock \href {https://doi.org/10.18653/v1/2025.blackboxnlp-1.5} {The
  comparative trap: Pairwise comparisons amplifies biased preferences of {LLM}
  evaluators}.
\newblock In \emph{Proceedings of the 8th BlackboxNLP Workshop: Analyzing and
  Interpreting Neural Networks for NLP}, pages 79--108, Suzhou, China.
  Association for Computational Linguistics.

\bibitem[{Luo et~al.(2024)Luo, Chen, He, and Sun}]{luo-etal-2024-prp}
Jian Luo, Xuanang Chen, Ben He, and Le~Sun. 2024.
\newblock \href {https://doi.org/10.18653/v1/2024.acl-long.313} {{PRP}-graph:
  Pairwise ranking prompting to {LLM}s with graph aggregation for effective
  text re-ranking}.
\newblock In \emph{Proceedings of the 62nd Annual Meeting of the Association
  for Computational Linguistics (Volume 1: Long Papers)}, pages 5766--5776,
  Bangkok, Thailand. Association for Computational Linguistics.

\bibitem[{Mohajer et~al.(2017)Mohajer, Suh, and Elmahdy}]{mohajer2017active}
Soheil Mohajer, Changho Suh, and Adel Elmahdy. 2017.
\newblock \href {https://proceedings.mlr.press/v70/mohajer17a.html} {Active
  learning for top-$k$ rank aggregation from noisy comparisons}.
\newblock In \emph{Proceedings of the 34th International Conference on Machine
  Learning}, volume~70 of \emph{Proceedings of Machine Learning Research},
  pages 2488--2497. PMLR.

\bibitem[{Qin et~al.(2024)Qin, Jagerman, Hui, Zhuang, Wu, Yan, Shen, Liu, Liu,
  Metzler, Wang, and Bendersky}]{qin2024pairwise}
Zhen Qin, Rolf Jagerman, Kai Hui, Honglei Zhuang, Junru Wu, Le~Yan, Jiaming
  Shen, Tianqi Liu, Jialu Liu, Donald Metzler, Xuanhui Wang, and Michael
  Bendersky. 2024.
\newblock \href {https://doi.org/10.18653/v1/2024.findings-naacl.97} {Large
  language models are effective text rankers with pairwise ranking prompting}.
\newblock In \emph{Findings of the Association for Computational Linguistics:
  NAACL 2024}, pages 1504--1518, Mexico City, Mexico. Association for
  Computational Linguistics.

\bibitem[{Ren et~al.(2020)Ren, Liu, and Shroff}]{ren2020bestk}
Wenbo Ren, Jia Liu, and Ness Shroff. 2020.
\newblock \href {https://proceedings.mlr.press/v119/ren20a.html} {The sample
  complexity of best-$k$ items selection from pairwise comparisons}.
\newblock In \emph{Proceedings of the 37th International Conference on Machine
  Learning}, volume 119 of \emph{Proceedings of Machine Learning Research},
  pages 8051--8072.

\bibitem[{Shah and Wainwright(2018)}]{shah2018simplerobust}
Nihar~B. Shah and Martin~J. Wainwright. 2018.
\newblock \href {https://jmlr.org/papers/v18/16-206.html} {Simple, robust and
  optimal ranking from pairwise comparisons}.
\newblock \emph{Journal of Machine Learning Research}, 18(199):1--38.

\bibitem[{Shi et~al.(2024)Shi, Ma, Liang, Diao, Ma, and
  Vosoughi}]{shi2024judges}
Lin Shi, Chiyu Ma, Wenhua Liang, Xingjian Diao, Weicheng Ma, and Soroush
  Vosoughi. 2024.
\newblock \href {https://doi.org/10.48550/arXiv.2406.07791} {Judging the
  judges: A systematic study of position bias in llm-as-a-judge}.
\newblock \emph{arXiv preprint arXiv:2406.07791}.

\bibitem[{Sun et~al.(2025)Sun, Zhong, Zhou, and Han}]{sun2025dynamicrag}
Jiashuo Sun, Xianrui Zhong, Sizhe Zhou, and Jiawei Han. 2025.
\newblock \href {https://doi.org/10.48550/arXiv.2505.07233} {Dynamicrag:
  Leveraging outputs of large language model as feedback for dynamic reranking
  in retrieval-augmented generation}.
\newblock \emph{arXiv preprint arXiv:2505.07233}.

\bibitem[{Sun et~al.(2023)Sun, Yan, Ma, Wang, Ren, Chen, Yin, and
  Ren}]{sun2023chatgptsearch}
Weiwei Sun, Lingyong Yan, Xinyu Ma, Shuaiqiang Wang, Pengjie Ren, Zhumin Chen,
  Dawei Yin, and Zhaochun Ren. 2023.
\newblock \href {https://doi.org/10.18653/v1/2023.emnlp-main.923} {Is
  {C}hat{GPT} good at search? investigating large language models as re-ranking
  agents}.
\newblock In \emph{Proceedings of the 2023 Conference on Empirical Methods in
  Natural Language Processing}, pages 14918--14937, Singapore. Association for
  Computational Linguistics.

\bibitem[{Wang et~al.(2025)Wang, Xia, Liao, Wang, and Liu}]{wang2025realm}
Pinhuan Wang, Zhiqiu Xia, Chunhua Liao, Feiyi Wang, and Hang Liu. 2025.
\newblock \href {https://doi.org/10.18653/v1/2025.emnlp-main.1218} {{REALM}:
  Recursive relevance modeling for {LLM}-based document re-ranking}.
\newblock In \emph{Proceedings of the 2025 Conference on Empirical Methods in
  Natural Language Processing}, pages 23875--23889, Suzhou, China. Association
  for Computational Linguistics.

\bibitem[{Wisznia et~al.(2025)Wisznia, Bola{\~n}os, Tollo, Marraffini,
  Gianolini, Hsueh, and Del~Corro}]{wisznia2025batchingcaching}
Juan Wisznia, Cecilia Bola{\~n}os, Juan Tollo, Giovanni Franco~Gabriel
  Marraffini, Agust{\'i}n~Andr{\'e}s Gianolini, Noe~Fabian Hsueh, and Luciano
  Del~Corro. 2025.
\newblock \href {https://doi.org/10.18653/v1/2025.acl-short.83} {Are optimal
  algorithms still optimal? rethinking sorting in {LLM}-based pairwise ranking
  with batching and caching}.
\newblock In \emph{Proceedings of the 63rd Annual Meeting of the Association
  for Computational Linguistics (Volume 2: Short Papers)}, pages 1064--1072,
  Vienna, Austria. Association for Computational Linguistics.

\bibitem[{Wu et~al.(2025)Wu, Shrivastava, Zhu, Samuel, Kumar, and
  Liu}]{wu2025realtimeprp}
Jingyu Wu, Aditya Shrivastava, Jing Zhu, Alfy Samuel, Anoop Kumar, and Daben
  Liu. 2025.
\newblock \href {https://doi.org/10.48550/arXiv.2511.07555} {Llm optimization
  unlocks real-time pairwise reranking}.
\newblock \emph{arXiv preprint arXiv:2511.07555}.

\bibitem[{Yin et~al.(2025)Yin, Vardi, and Choudhary}]{yin2025fragile}
Haonan Yin, Shai Vardi, and Vidyanand Choudhary. 2025.
\newblock \href {https://doi.org/10.48550/arXiv.2506.14092} {Fragile
  preferences: A deep dive into order effects in large language models}.
\newblock \emph{arXiv preprint arXiv:2506.14092}.

\bibitem[{Zhou et~al.(2025)Zhou, Luo, Feng, and Wang}]{zhou2025large}
Yinxin Zhou, Qin Luo, Bin Feng, and Bang Wang. 2025.
\newblock \href {https://doi.org/10.36227/techrxiv.176300630.01740917/v1}
  {Large language models for reranking: A survey}.
\newblock \url{https://doi.org/10.36227/techrxiv.176300630.01740917/v1}.
\newblock TechRxiv preprint.

\bibitem[{Zhu et~al.(2023)Zhu, Yuan, Wang, Liu, Liu, Deng, Chen, Liu, Dou, and
  Wen}]{zhu2023llmir}
Yutao Zhu, Huaying Yuan, Shuting Wang, Jiongnan Liu, Wenhan Liu, Chenlong Deng,
  Haonan Chen, Zheng Liu, Zhicheng Dou, and Ji-Rong Wen. 2023.
\newblock \href {https://doi.org/10.48550/arXiv.2308.07107} {Large language
  models for information retrieval: A survey}.
\newblock \emph{arXiv preprint arXiv:2308.07107}.

\bibitem[{Zhuang et~al.(2024)Zhuang, Zhuang, Koopman, and
  Zuccon}]{zhuang2024setwise}
Shengyao Zhuang, Honglei Zhuang, Bevan Koopman, and Guido Zuccon. 2024.
\newblock \href {https://doi.org/10.1145/3626772.3657813} {A setwise approach
  for effective and highly efficient zero-shot ranking with large language
  models}.
\newblock In \emph{Proceedings of the 47th International ACM SIGIR Conference
  on Research and Development in Information Retrieval (SIGIR '24)}, pages
  38--47. ACM.

\end{thebibliography}


\FloatBarrier
\appendix
\section{Parallelization Opportunities}
\label{sec:parallelization}

Both our proposed active ranking algorithms exhibit substantial parallelization potential that could significantly reduce wall-clock latency in practice.

\paragraph{Mohajer.}
The Mohajer algorithm's structure naturally lends itself to parallel execution at all levels.
First, the $K$ independent tournaments used for champion selection can be executed concurrently, as each tournament operates on a disjoint subset of candidates with no inter-group dependencies.
Second, the heapify operation itself is amenable to parallel construction techniques, allowing the initial heap formation to proceed in $O(\log n)$ depth rather than sequential $O(n)$ time.
Finally, within each tournament, pairwise comparisons at the same tree depth are independent and can be batched for simultaneous LLM inference.

\paragraph{Optimized PAC.}
Our optimized PAC algorithm also presents parallelization opportunities, where comparisons of candidates against the selected anchors are entirely independent.
With $K/2$ anchors and a candidate pool of size $K \times \text{pool\_mult}$, these $O(K^2)$ comparisons can be executed in parallel, subject only to the LLM inference throughput.
The subsequent winner-set construction and greedy accumulation require only lightweight aggregation, which remains negligible compared to comparison costs.

\paragraph{Latency implications.}
Assuming an LLM inference API with high throughput (e.g., batched GPU inference or distributed serving), parallelization could theoretically reduce Mohajer's latency from $O(n \log K)$ sequential comparisons to $O(\log Q \cdot \log K)$ parallel rounds, and PAC's latency from $O(K \sqrt{n})$ comparisons to $O(\sqrt{n})$ rounds with batch size $K$.
In our TREC DL experiments, where Mohajer performs $\sim$350 comparisons and PAC $\sim$185 comparisons, parallelization with batch size 10 could reduce round counts to $\sim$35 and $\sim$19 respectively, yielding substantial speedups on systems where comparison latency dominates over LLM throughput.

\section{Supplementary Tables}
\label{sec:appendix}
\FloatBarrier

\FloatBarrier

\noindent This section provides supplementary results and ablations referenced in the main text.
\vspace{0.5em}

\begingroup

\setcounter{topnumber}{5}
\setcounter{bottomnumber}{5}
\setcounter{totalnumber}{10}
\renewcommand{\topfraction}{0.95}
\renewcommand{\bottomfraction}{0.95}
\renewcommand{\textfraction}{0.05}
\renewcommand{\floatpagefraction}{0.95}

\setlength{\textfloatsep}{10pt plus 2pt minus 2pt}
\setlength{\floatsep}{8pt plus 2pt minus 2pt}
\setlength{\intextsep}{8pt plus 2pt minus 2pt}
\renewcommand{\arraystretch}{1.08}
\setlength{\tabcolsep}{3.5pt}

\captionsetup[table]{
  font=small,
  labelfont=bf,
  skip=3pt,
  justification=raggedright,
  singlelinecheck=false
}
\captionsetup[subtable]{
  font=small,
  justification=centering,
  singlelinecheck=false
}

\setcounter{table}{0}
\renewcommand{\thetable}{A.\arabic{table}}

\providecommand{\best}[1]{\textbf{#1}}
\providecommand{\second}[1]{\underline{#1}}
\providecommand{\conv}[1]{#1\textsuperscript{\textdagger}}
\providecommand{\postconv}[1]{\textcolor{black!55}{#1}}

\begin{table*}[!t]
\centering

\begin{subtable}{\textwidth}
\centering
\caption{TREC DL2019}
\label{tab:budget_ndcg10_oracles_flan_xl_dl2019}
\begin{adjustbox}{max width=\textwidth}
\scriptsize
\begin{tabular}{c l rrrrrrrrr}
\toprule
& & \multicolumn{9}{c}{Number of LLM calls (budget)} \\
\cmidrule(lr){3-11}
Oracle & Ranker & 100 & 150 & 200 & 250 & 300 & 350 & 400 & 450 & 500 \\
\midrule
\multirow{6}{*}{\rotatebox[origin=c]{90}{Bidirectional}} & BubbleSort & \underline{50.58} & \underline{50.58} & 57.13 & 57.13 & 57.13 & 57.78 & 60.64 & 60.69 & 60.95 \\
& HeapSort & 9.14 & 8.49 & 8.61 & 11.91 & 23.97 & 41.68 & 54.14 & 63.58 & \textbf{68.92} \\
& QuickSort & \textbf{58.08} & \textbf{58.08} & \underline{58.30} & \underline{58.57} & 58.57 & 58.57 & 58.85 & 59.08 & 59.27 \\
& PAC + Bubble & \underline{50.58} & \underline{50.58} & 50.58 & 50.58 & \underline{60.00} & \underline{62.59} & \underline{62.60} & 62.60 & 62.60 \\
& Mohajer + Bubble & 32.76 & 32.76 & \textbf{62.44} & \textbf{65.35} & \textbf{66.34} & \textbf{66.23} & \textbf{66.64} & \textbf{66.59} & \underline{66.59} \\
& Mohajer & 32.76 & 32.76 & \textbf{62.44} & \textbf{65.35} & \textbf{66.34} & \textbf{66.23} & \textbf{66.64} & \underline{66.58} & 66.58 \\
\midrule
\multirow{6}{*}{\rotatebox[origin=c]{90}{Randomized}} & BubbleSort & 55.92 & 56.88 & 59.77 & 60.33 & 61.98 & 62.52 & 64.27 & 64.37 & 65.73 \\
& HeapSort & 9.11 & 16.89 & 49.27 & 65.93 & 68.49 & \underline{69.03} & \textbf{69.82} & \underline{68.76} & \underline{68.84} \\
& QuickSort & \underline{56.33} & 57.57 & 56.98 & 59.89 & 58.42 & 61.44 & 63.00 & 64.83 & 64.37 \\
& PAC + Bubble & 50.58 & \underline{58.35} & \underline{61.05} & 61.05 & 61.05 & 61.05 & 61.05 & 61.05 & 61.05 \\
& Mohajer + Bubble & \textbf{61.41} & \textbf{67.79} & \textbf{68.70} & \underline{67.99} & \underline{68.53} & \textbf{69.17} & \underline{69.47} & \textbf{69.47} & \textbf{69.47} \\
& Mohajer & \textbf{61.41} & \textbf{67.79} & \textbf{68.70} & \textbf{68.73} & \textbf{68.73} & 68.73 & 68.73 & 68.73 & 68.73 \\
\bottomrule
\end{tabular}
\end{adjustbox}
\end{subtable}

\vspace{0.35em}

\begin{subtable}{\textwidth}
\centering
\caption{TREC DL2020}
\label{tab:budget_ndcg10_oracles_flan_xl_dl2020}
\begin{adjustbox}{max width=\textwidth}
\scriptsize
\begin{tabular}{c l rrrrrrrrr}
\toprule
& & \multicolumn{9}{c}{Number of LLM calls (budget)} \\
\cmidrule(lr){3-11}
Oracle & Ranker & 100 & 150 & 200 & 250 & 300 & 350 & 400 & 450 & 500 \\
\midrule
\multirow{6}{*}{\rotatebox[origin=c]{90}{Bidirectional}} & BubbleSort & \underline{47.96} & \underline{47.96} & \underline{55.72} & \underline{55.72} & \underline{55.72} & 56.18 & 59.86 & 59.90 & 60.07 \\
& HeapSort & 4.48 & 3.77 & 3.65 & 6.65 & 22.12 & 42.00 & 54.44 & 62.05 & \textbf{67.50} \\
& QuickSort & \textbf{53.77} & \textbf{53.69} & 53.44 & 53.83 & 53.83 & 53.83 & 53.95 & 54.10 & 54.10 \\
& PAC + Bubble & \underline{47.96} & \underline{47.96} & 47.96 & 47.96 & 55.04 & \underline{58.59} & \underline{58.62} & 58.62 & 58.62 \\
& Mohajer + Bubble & 27.48 & 27.48 & \textbf{62.24} & \textbf{64.25} & \textbf{65.83} & \textbf{66.32} & \textbf{67.02} & \textbf{67.45} & \underline{67.45} \\
& Mohajer & 27.48 & 27.48 & \textbf{62.24} & \textbf{64.25} & \textbf{65.83} & \textbf{66.32} & \underline{66.98} & \underline{67.35} & 67.35 \\
\midrule
\multirow{6}{*}{\rotatebox[origin=c]{90}{Randomized}} & BubbleSort & \underline{55.23} & \underline{55.31} & \underline{59.03} & 58.45 & 61.30 & 61.78 & 63.35 & 64.37 & 65.88 \\
& HeapSort & 4.59 & 15.96 & 51.38 & 64.88 & \underline{67.31} & \textbf{67.64} & \underline{67.27} & \textbf{67.78} & \textbf{68.58} \\
& QuickSort & 51.51 & 53.28 & 54.34 & 54.67 & 56.15 & 56.05 & 57.25 & 59.76 & 61.91 \\
& PAC + Bubble & 47.96 & 55.27 & 58.51 & 58.51 & 58.51 & 58.51 & 58.51 & 58.51 & 58.51 \\
& Mohajer + Bubble & \textbf{59.95} & \textbf{65.06} & \textbf{67.66} & \underline{67.61} & 67.05 & 67.60 & 66.30 & 66.30 & 66.30 \\
& Mohajer & \textbf{59.95} & \textbf{65.06} & \textbf{67.66} & \textbf{67.62} & \textbf{67.62} & \underline{67.62} & \textbf{67.62} & \underline{67.62} & \underline{67.62} \\
\bottomrule
\end{tabular}
\end{adjustbox}
\end{subtable}

\caption{
Budgeted NDCG@10 (\%) on TREC DL2019 and DL2020 using Flan-T5-XL under bidirectional vs.\ randomized-direction prompting.
Budgets denote the number of LLM calls. Within each oracle block, bold/underline indicate best/second-best per budget column among the listed rankers.
}
\label{tab:budget_ndcg10_oracles_flan_xl_by_dataset}
\end{table*}
\begin{table*}[!t]
\centering
\small
\begin{adjustbox}{max width=\textwidth}
\begin{tabular}{c l rrrrrrrrr}
\toprule
& & \multicolumn{9}{c}{Number of LLM calls (budget)} \\
\cmidrule(lr){3-11}
$k$ & Ranker & 100 & 150 & 200 & 250 & 300 & 350 & 400 & 450 & 500 \\
\midrule
\multirow{2}{*}{10} & BubbleSort
& 55.57 & 56.10 & 59.40 & 59.39 & 61.64 & 62.15 & 63.81 & 64.37 & 65.80 \\
& Mohajer
& \best{60.68} & \best{66.42} & \conv{\best{68.18}} & \postconv{\best{68.18}} & \postconv{\best{68.18}} & \postconv{\best{68.18}} & \postconv{\best{68.18}} & \postconv{\best{68.18}} & \postconv{\best{68.18}} \\
\midrule
\multirow{2}{*}{20} & BubbleSort
& \best{52.73} & 53.07 & 55.76 & 55.62 & 57.36 & 57.60 & 59.18 & 59.31 & 60.49 \\
& Mohajer
& 51.51 & \best{58.30} & \best{61.82} & \conv{\best{62.98}} & \postconv{\best{62.98}} & \postconv{\best{62.98}} & \postconv{\best{62.98}} & \postconv{\best{62.98}} & \postconv{\best{62.98}} \\
\midrule
\multirow{2}{*}{30} & BubbleSort
& \best{51.62} & 52.02 & 54.06 & 53.88 & 55.49 & 55.61 & 57.07 & 57.19 & 58.02 \\
& Mohajer
& 48.61 & \best{54.04} & \best{56.69} & \conv{\best{58.22}} & \postconv{\best{58.22}} & \postconv{\best{58.22}} & \postconv{\best{58.22}} & \postconv{\best{58.22}} & \postconv{\best{58.22}} \\
\midrule
\multirow{2}{*}{40} & BubbleSort
& \best{51.47} & 51.78 & 53.77 & 53.58 & 55.17 & 55.44 & \best{56.38} & \best{56.86} & \best{57.19} \\
& Mohajer
& 48.06 & \best{52.94} & \best{54.49} & \conv{\best{56.03}} & \postconv{\best{56.03}} & \postconv{\best{56.03}} & \postconv{56.03} & \postconv{56.03} & \postconv{56.03} \\
\midrule
\multirow{2}{*}{50} & BubbleSort
& \best{51.46} & 52.38 & 53.59 & 53.83 & 54.71 & \best{55.31} & \best{55.80} & \best{56.17} & \best{56.49} \\
& Mohajer
& 48.12 & \best{52.72} & \best{53.96} & \conv{\best{55.03}} & \postconv{\best{55.03}} & \postconv{55.03} & \postconv{55.03} & \postconv{55.03} & \postconv{55.03} \\
\bottomrule
\end{tabular}
\end{adjustbox}
\caption{
Top-$k$ sweep with randomized-direction prompting (Flan-T5-XL). Values are average NDCG@$k$ (\%) over TREC DL2019 and DL2020.
Budgets denote the number of LLM calls. For each $(k,\text{budget})$, bold marks the better of BubbleSort vs.\ Mohajer.
\textsuperscript{\textdagger} indicates the first budget where the method is treated as converged; subsequent budgets are de-emphasized.
}
\label{tab:k_budget_sweep_randomized}
\par\vspace{2pt}\footnotesize\noindent\textit{Note.} Mohajer dominates BubbleSort when $k$ and the budget are low. As $k$ increases, the crossover point where BubbleSort overtakes Mohajer arrives sooner in terms of number of LLM calls.
\end{table*}
\begin{table*}[!t]
\centering
\small
\begin{adjustbox}{max width=\textwidth}
\begin{tabular}{@{}llcccc@{}}
\toprule
\multirow{2}{*}{LLM} & \multirow{2}{*}{Method}
& \multicolumn{2}{c}{DL2019}
& \multicolumn{2}{c}{DL2020} \\
\cmidrule(lr){3-4}\cmidrule(lr){5-6}
& & NDCG@10 (\%) & \#Inf. & NDCG@10 (\%) & \#Inf. \\
\midrule
NA & BM25 & 50.6 & -- & 48.0 & -- \\
\midrule
\multirow{3}{*}{Flan-T5-L}
& PRP-Graph-10     & \best{65.8}    & 492.7  & \best{61.8}    & 492.5  \\
& Mohajer + Bu. & \second{61.75} & 354.60 & \second{58.8}  & 354.245 \\
& PAC + Bu.     & 59.9           & 183.34 & 57.29          & 183.99 \\
\midrule
\multirow{3}{*}{Flan-T5-XL}
& PRP-Graph-10     & \second{67.6}  & 492.6  & \second{66.1}  & 492.5 \\
& Mohajer + Bu. & \best{69.47}   & 338.11 & \best{66.3}    & 338.74 \\
& PAC + Bu.     & 61.05          & 184.16 & 58.51          & 183.72 \\
\midrule
\multirow{3}{*}{Flan-T5-XXL}
& PRP-Graph-10     & \second{66.6}  & 492.6  & \second{66.1}  & 492.6 \\
& Mohajer + Bu. & \best{68.54}   & 341.81 & \best{68.96}   & 344.17 \\
& PAC + Bu.     & 60.92          & 183.46 & 60.71          & 183.59 \\
\bottomrule
\end{tabular}
\end{adjustbox}
\caption{
Comparison to PRP-Graph on TREC DL2019/2020. NDCG@10 is reported in percent.
``\#Inf.'' is the average number of inferred pairwise comparisons used by the method (lower is fewer comparisons).
Bold/underline indicate best/second-best NDCG@10 within each (DL year, LLM) block.
}
\label{tab:trec_prpgraph_comp}
\par\vspace{2pt}\footnotesize\noindent\textit{Note.} Our results show that with larger Flan models, Mohajer + Bubble has increasingly better results than PRP-Graph with fewer comparisons. PAC + Bubble falls behind in performance but has significantly fewer comparisons. 
\end{table*}
\begin{table*}[!t]
\centering
\footnotesize
\setlength{\tabcolsep}{3pt}
\renewcommand{\arraystretch}{1.12}
\begin{adjustbox}{max width=\textwidth}
\begin{tabular}{llcccccccccc}
\toprule
\multicolumn{2}{c}{\textbf{Method}} &
\multicolumn{8}{c}{\textbf{Dataset (NDCG@10 \%)}} &
\textbf{Avg. NDCG@10} & \textbf{Avg. Calls/Task} \\
\cmidrule(lr){1-2}\cmidrule(lr){3-10}\cmidrule(lr){11-12}
\textbf{Reranker} & \textbf{Oracle} &
\textbf{Covid} & \textbf{Robust04} & \textbf{Touche} & \textbf{SciFact} &
\textbf{DBPedia} & \textbf{DL19} & \textbf{DL20} & \textbf{FiQA} & & \\
\midrule
\multirow{2}{*}{BubbleSort@10} & Bidirectional
& 70.9 & 44.2 & \textbf{44.7} & \textbf{69.2} & 41.7 & 63.4 & 58.6 & 29.4 & 52.8 & \textbf{653} \\
& Randomized
& \textbf{75.7} & \textbf{46.3} & 41.3 & 67.7 & \textbf{43.6} & \textbf{67.0} & \textbf{62.5} & \textbf{32.4} & \textbf{54.6} & 930 \\
\midrule
\multirow{2}{*}{HeapSort} & Bidirectional
& 76.0 & \textbf{40.4} & \textbf{33.2} & \textbf{67.5} & \textbf{41.4} & \textbf{65.0} & \textbf{62.6} & \textbf{31.3} & \textbf{52.7} & 1230 \\
& Randomized
& \textbf{76.9} & 38.4 & 27.7 & 63.3 & 39.8 & 62.6 & 58.5 & 27.7 & 49.0 & \textbf{860} \\
\midrule
\multirow{2}{*}{QuickSort} & Bidirectional
& \textbf{76.2} & \textbf{41.0} & \textbf{27.4} & \textbf{60.1} & \textbf{41.1} & \textbf{64.5} & \textbf{58.5} & \textbf{26.8} & \textbf{49.5} & 1954 \\
& Randomized
& 75.8 & 35.4 & 25.1 & 58.7 & 39.1 & 61.8 & 58.1 & 24.6 & 47.3 & \textbf{556} \\
\midrule
\multirow{2}{*}{Mohajer} & Bidirectional
& \textbf{76.5} & \textbf{37.5} & \textbf{26.4} & 53.8 & \textbf{39.7} & \textbf{62.6} & 56.1 & \textbf{25.9} & \textbf{47.3} & 399 \\
& Randomized
& 76.2 & 36.2 & 24.4 & \textbf{57.5} & 39.1 & 60.6 & \textbf{57.2} & 24.9 & 47.0 & \textbf{232} \\
\midrule
\multirow{2}{*}{Mohajer+Bubble} & Bidirectional
& 76.5 & 37.5 & \textbf{26.4} & 53.9 & 39.7 & \textbf{62.6} & 56.1 & 25.9 & 47.3 & 423 \\
& Randomized
& \textbf{76.9} & \textbf{37.8} & 25.7 & \textbf{58.8} & \textbf{40.0} & 61.8 & \textbf{58.8} & \textbf{26.2} & \textbf{48.2} & \textbf{354} \\
\midrule
\multirow{2}{*}{Pac+Bubble} & Bidirectional
& 69.3 & \textbf{44.0} & \textbf{41.4} & \textbf{68.5} & \textbf{39.2} & \textbf{61.7} & 57.2 & \textbf{29.2} & \textbf{51.3} & 323 \\
& Randomized
& \textbf{70.2} & 41.0 & 38.2 & 67.0 & 38.1 & 60.0 & \textbf{57.3} & 27.2 & 49.9 & \textbf{184} \\
\bottomrule
\end{tabular}
\end{adjustbox}
\caption{
End-to-end NDCG@10 (\%) and average pairwise-comparison calls per task. For each reranker and column, \textbf{bold} indicates the better oracle variant (higher NDCG; lower calls).
If the two oracle variants tie after rounding, both are bolded.
}
\label{tab:reranker_oracle_compare}
\par\vspace{2pt}\footnotesize\noindent\textit{Note.} Active-learning rankers are more robust to noisy comparisons and profit most from randomized-direction sampling. BubbleSort requires many more comparisons as cycles emerge, but said cycles allow the latent transitive ordering to surface more than with a deterministic oracle, yielding higher NDCG@10.

\end{table*}

\begin{table*}[!hbp]
\centering
\small
\renewcommand{\arraystretch}{1.15}
\setlength{\tabcolsep}{6pt}

\begin{minipage}[t]{0.48\textwidth}
\centering
\textbf{Stratified by dataset}\\[2pt]
\begin{tabular}{lrrr}
\hline
Dataset & Flip-rate & \#Pairs & \#Flips \\
\hline
All   & 20.62\% & 1,202,850 & 248,021 \\
DL19  & 21.40\% &   212,850 &  45,542 \\
DL20  & 20.45\% &   990,000 & 202,479 \\
\hline
\end{tabular}
\end{minipage}\hfill
\begin{minipage}[t]{0.48\textwidth}
\centering
\textbf{Stratified by BM25 rank distance $|r(i)-r(j)|$}\\[2pt]
\begin{tabular}{lrrr}
\hline
$|r(i)-r(j)|$ & Flip-rate & \#Pairs & \#Flips \\
\hline
1--5    & 22.56\% &  92,639 &  20,903 \\
6--10   & 20.73\% &  86,645 &  17,965 \\
11--20  & 20.85\% & 160,762 &  33,524 \\
$>20$   & 20.36\% & 862,804 & 175,629 \\
\hline
\end{tabular}
\end{minipage}

\caption{Bidirectional flip-rate sanity check, Flan-T5-XL (order effects). For each candidate pair $\{d_i,d_j\}$, we query the judge twice (once as $(d_i,d_j)$ and once as $(d_j,d_i)$) and report the fraction of pairs whose winner flips. Results are shown overall and stratified by dataset and BM25 rank distance.}
\label{tab:flip_rate_order_effects}
\end{table*}

\begin{table*}[!hbp]
\centering
\begin{adjustbox}{max width=\textwidth}
\scriptsize
\begin{tabular}{c l rrrrrrrrr}
\toprule
& & \multicolumn{9}{c}{Number of LLM calls (budget)} \\
\cmidrule(lr){3-11}
Oracle & Ranker & 100 & 150 & 200 & 250 & 300 & 350 & 400 & 450 & 500 \\
\midrule

\multirow{6}{*}{\rotatebox[origin=c]{90}{Bidirectional}}
& BubbleSort        & \underline{50.58} & \underline{50.58} & 57.04 & 57.04 & 57.18 & 58.73 & 60.98 & 61.30 & 61.90 \\
& HeapSort          & 8.64 & 8.16 & 7.99 & 10.48 & 25.65 & 42.42 & 55.85 & 64.24 & 69.14 \\
& QuickSort         & \textbf{58.28} & \textbf{58.11} & \underline{58.04} & \underline{58.63} & 58.71 & 58.71 & 58.71 & 59.09 & 59.67 \\
& PAC + Bubble      & \underline{50.58} & \underline{50.58} & 50.58 & 50.58 & \underline{60.74} & \conv{\underline{64.25}} & \postconv{\underline{64.25}} & \postconv{64.25} & \postconv{64.25} \\
& Mohajer + Bubble  & 32.76 & 32.76 & \textbf{62.97} & \textbf{66.69} & \textbf{68.60} & \textbf{69.39} & \textbf{70.08} & \conv{\textbf{70.49}} & \postconv{\textbf{70.49}} \\
& Mohajer           & 32.76 & 32.76 & \textbf{62.97} & \textbf{66.69} & \textbf{68.60} & \textbf{69.39} & \textbf{70.08} & \conv{\underline{70.43}} & \postconv{\underline{70.43}} \\

\midrule

\multirow{6}{*}{\rotatebox[origin=c]{90}{Randomized}}
& BubbleSort        & \underline{57.49} & \underline{57.79} & 61.11 & 61.45 & 64.32 & 64.24 & 66.44 & 66.27 & 67.12 \\
& HeapSort          & 8.20 & 16.13 & 53.18 & 68.15 & \textbf{71.52} & 69.61 & \underline{70.73} & \textbf{71.14} & \underline{70.70} \\
& QuickSort         & 54.94 & 56.56 & 58.25 & 57.03 & 59.96 & 60.92 & 62.13 & 61.92 & 64.81 \\
& PAC + Bubble      & 50.58 & 57.00 & \conv{\underline{63.02}} & \postconv{63.02} & \postconv{63.02} & \postconv{63.02} & \postconv{63.02} & \postconv{63.02} & \postconv{63.02} \\
& Mohajer + Bubble  & \textbf{63.37} & \textbf{66.49} & \textbf{69.37} & \underline{68.82} & 70.33 & \underline{70.44} & \conv{70.26} & \postconv{70.26} & \postconv{70.26} \\
& Mohajer           & \textbf{63.37} & \textbf{66.49} & \textbf{69.37} & \conv{\textbf{70.98}} & \postconv{\underline{70.98}} & \postconv{\textbf{70.98}} & \postconv{\textbf{70.98}} & \postconv{\underline{70.98}} & \postconv{\textbf{70.98}} \\

\bottomrule
\end{tabular}%
\end{adjustbox}
\caption{Average NDCG@10 (\%) on TREC DL 2019 with Qwen3-4B-Instruct across comparison budgets. Bold = best per column; underline = second-best per column (within each oracle block). \textsuperscript{\textdagger} indicates the smallest budget at which a method completes; results at larger budgets are visually de-emphasized.}
\label{tab:budget_ndcg10_oracles_qwen3_4b_dl2019}
\par\vspace{2pt}\footnotesize\noindent\textit{Note.} Though not always the best in performance, active ranking algorithms reach competitive NDCG@10 relative to the classic algorithms, with drastic improvements in terms of average calls per task.

\end{table*}

\begin{table*}[!hbp]
\centering
\scriptsize
\setlength{\tabcolsep}{2.0pt}
\renewcommand{\arraystretch}{1.00}
\resizebox{0.90\textwidth}{!}{%
\begin{tabular}{llrrrr}
\toprule
\multicolumn{2}{c}{\textbf{Method}} &
\multicolumn{2}{c}{\textbf{Dataset (NDCG@10 \%)}} &
\textbf{Avg. NDCG@10} & \textbf{Avg. Calls/Task} \\
\cmidrule(lr){1-2}\cmidrule(lr){3-4}\cmidrule(lr){5-6}
& \textbf{Ranker (Oracle)} & \textbf{DL19} & \textbf{DL20} & & \\
\midrule
& \textbf{BM25} & 50.60 & 48.00 & 49.30 & -- \\
\midrule

\multirow{9}{*}{\rotatebox{90}{\textbf{Qwen3-4B-Instruct}}}
& BubbleSort (Bidirectional) & 67.44 & \textbf{70.45} & 68.95 & 1062.69 \\
& HeapSort (Bidirectional)   & 70.26 & \underline{69.46} & \textbf{69.86} & 1383.02 \\
& QuickSort (Bidirectional)  & \underline{70.70} & 68.83 & \underline{69.77} & 1567.02 \\
\arrayrulecolor{black!35}
\cmidrule(lr){2-6}
\arrayrulecolor{black}
& PAC + Bubble (Bidirectional) & 64.25 & 67.09 & 65.67 & 332.55 \\
& PAC + Bubble (Randomized)    & 63.02 & 66.05 & 64.54 & \textbf{183.42} \\
& Mohajer + Bubble (Bidirectional) & 70.49 & 66.92 & 68.70 & 425.67 \\
& Mohajer + Bubble (Randomized)    & 70.26 & 67.58 & 68.92 & 346.54 \\
& Mohajer (Bidirectional)          & 70.43 & 66.92 & 68.67 & 398.26 \\
& Mohajer (Randomized)             & \textbf{70.98} & 65.43 & 68.20 & \underline{232.01} \\
\bottomrule
\end{tabular}%
}
\caption{
TREC DL2019/2020 end-to-end NDCG@10 (\%) and average pairwise-comparison calls per task using \textbf{Qwen3-4B-Instruct}.
Baselines are reported with a bidirectional oracle; our methods (PAC+Bubble and Mohajer variants) include both bidirectional and randomized comparisons.
Within the Qwen block, best is bold and second-best is underlined per column (higher NDCG; lower calls).
}
\label{tab:qwen3_4b_end2end}

\par\vspace{2pt}\footnotesize\noindent\textit{Note.} Though not always the best in performance, active ranking algorithms reach competitive NDCG@10 relative to the classic algorithms, with drastic improvements in terms of average calls per task.

\end{table*}

\endgroup  

\clearpage
\section{Supplementary Graphs}
\noindent This section presents supplementary figures for the latency experiments referenced in the main text.

\begin{figure}[H]
    \centering
    \includegraphics[width=\linewidth]{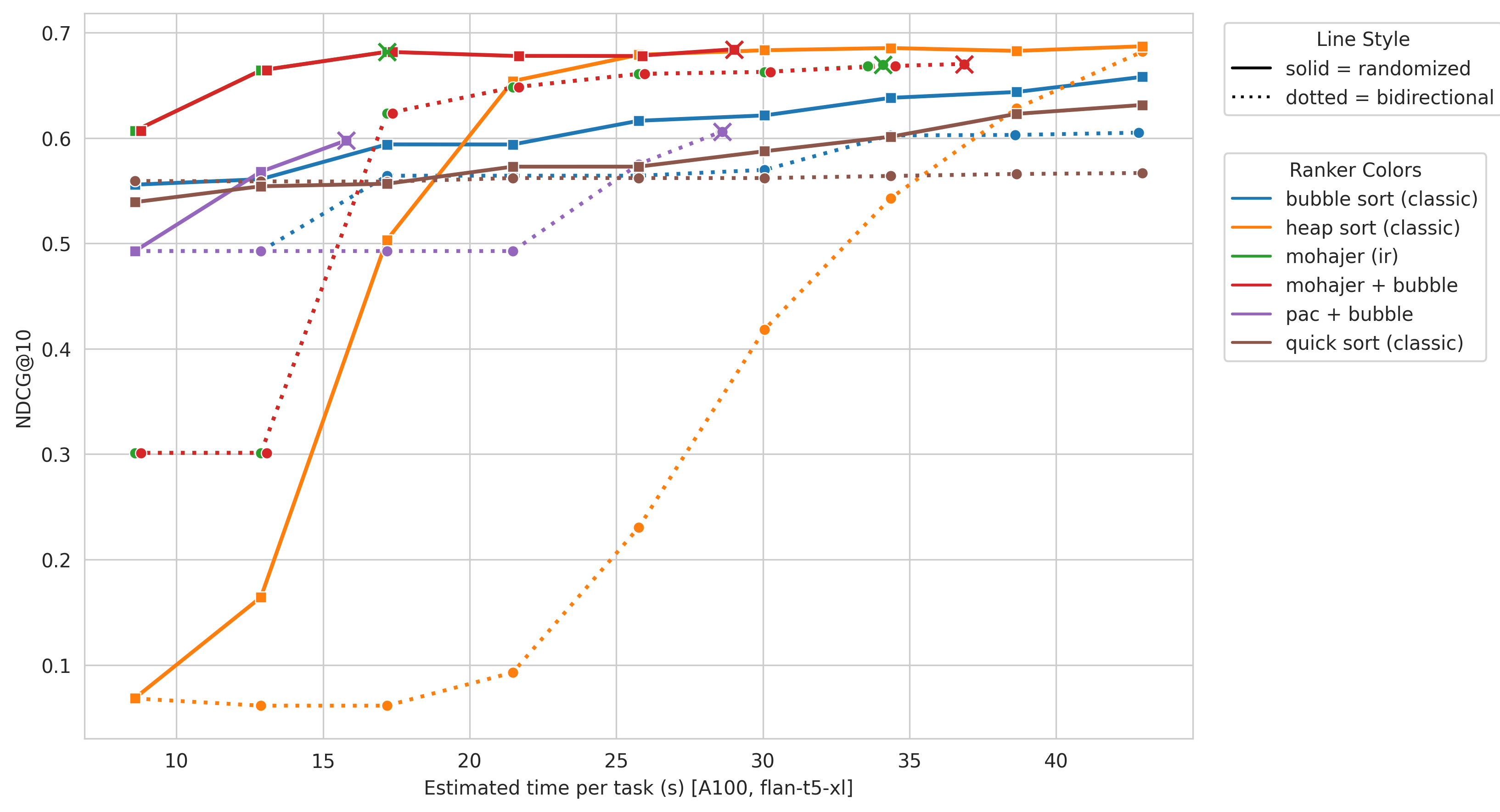}\par
    \vspace{0.6em}
    \includegraphics[width=\linewidth]{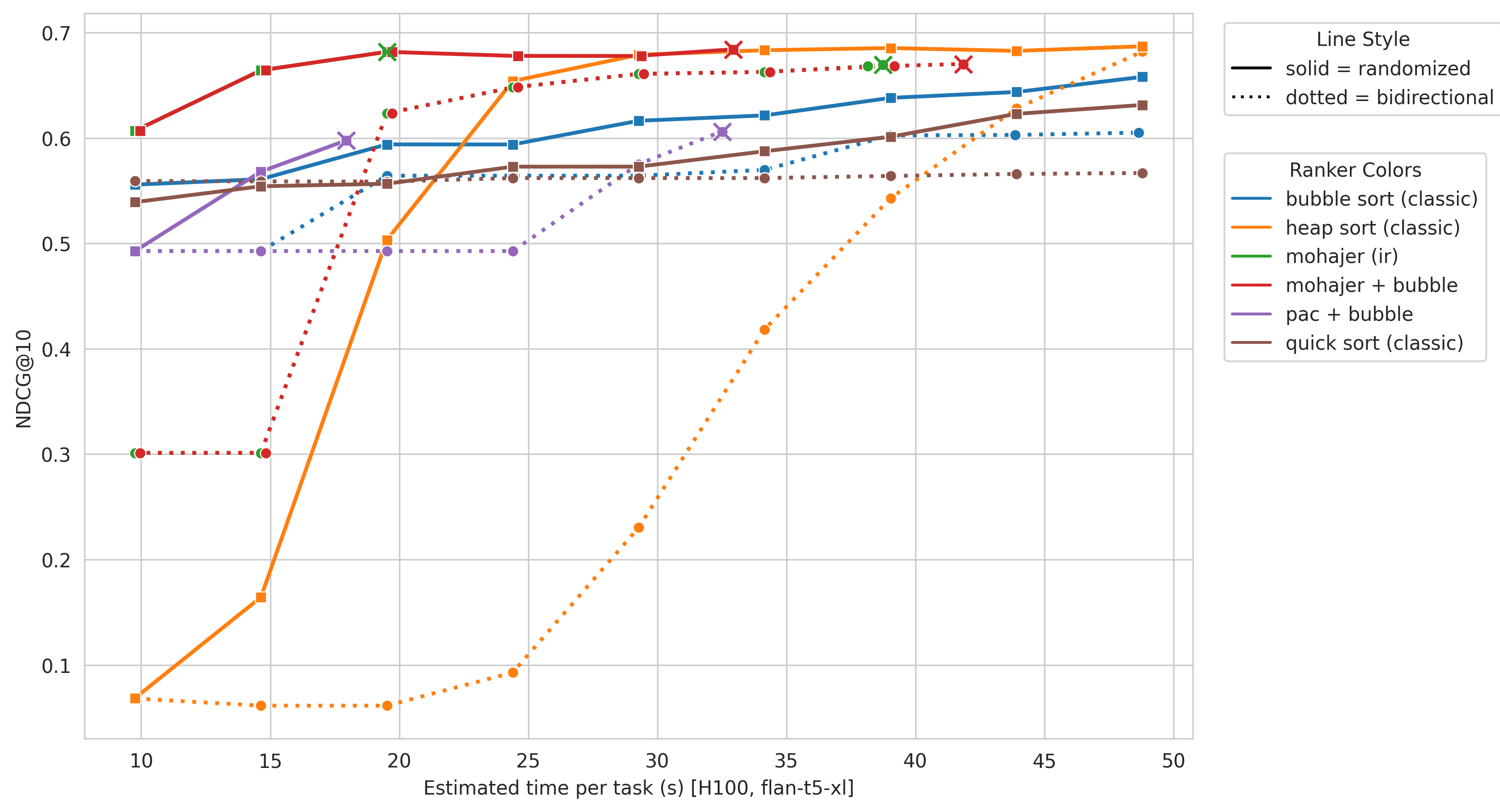}\par
    \vspace{0.6em}
    \includegraphics[width=\linewidth]{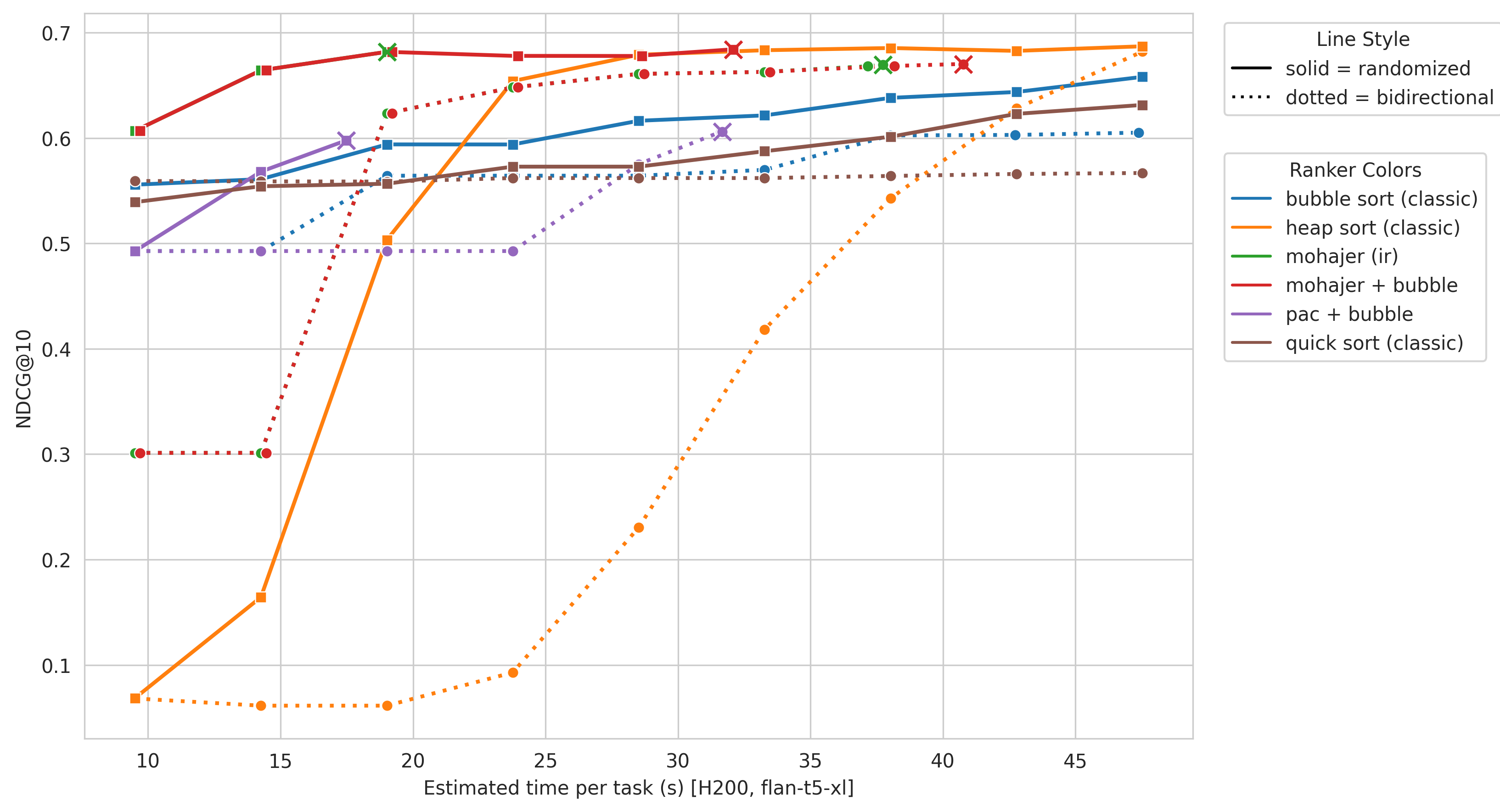}
    \caption{TREC DL 2019 and DL 2020 (Flan-T5-XL): NDCG@10 vs estimated time per task across GPUs, with both oracles shown. Colors denote rankers; solid lines are randomized and dotted lines are bidirectional oracles. X marks show when an algorithm has converged.}
    \label{fig:limit_time_all_flan_all_gpus}
\end{figure}

\begin{figure}[H]
    \centering
    \includegraphics[width=\linewidth]{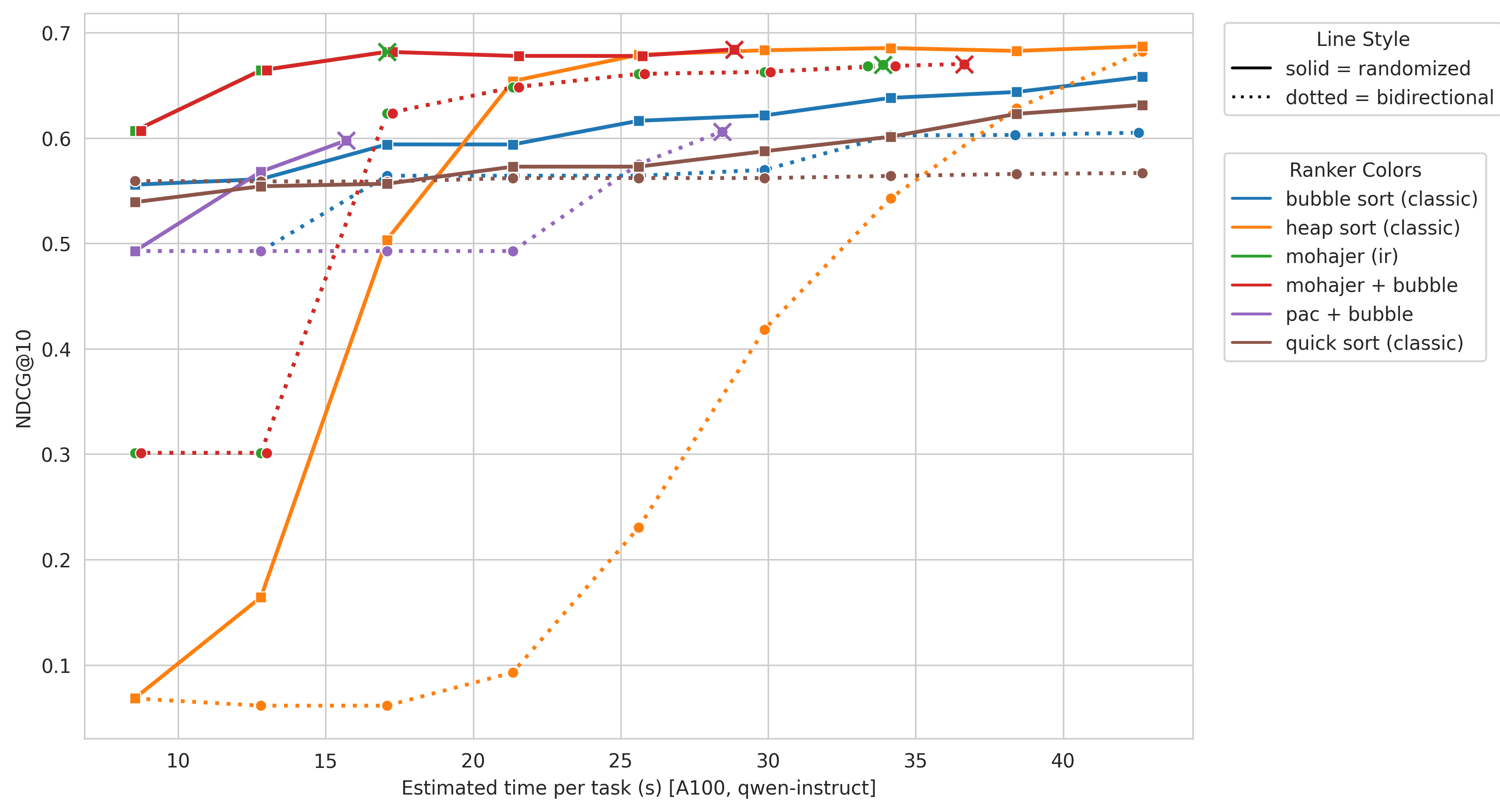}\par
    \vspace{0.6em}
    \includegraphics[width=\linewidth]{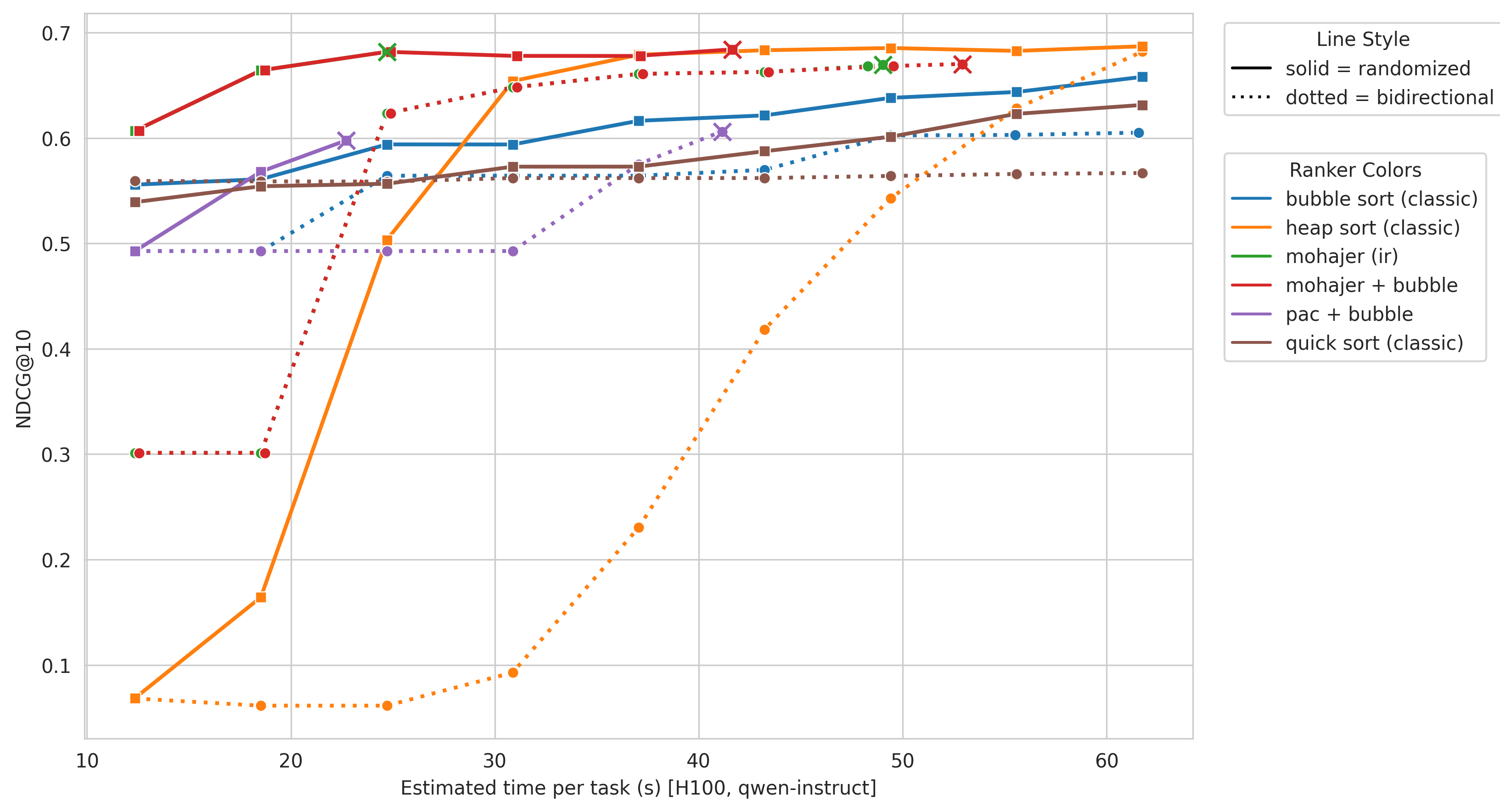}\par
    \vspace{0.6em}
    \includegraphics[width=\linewidth]{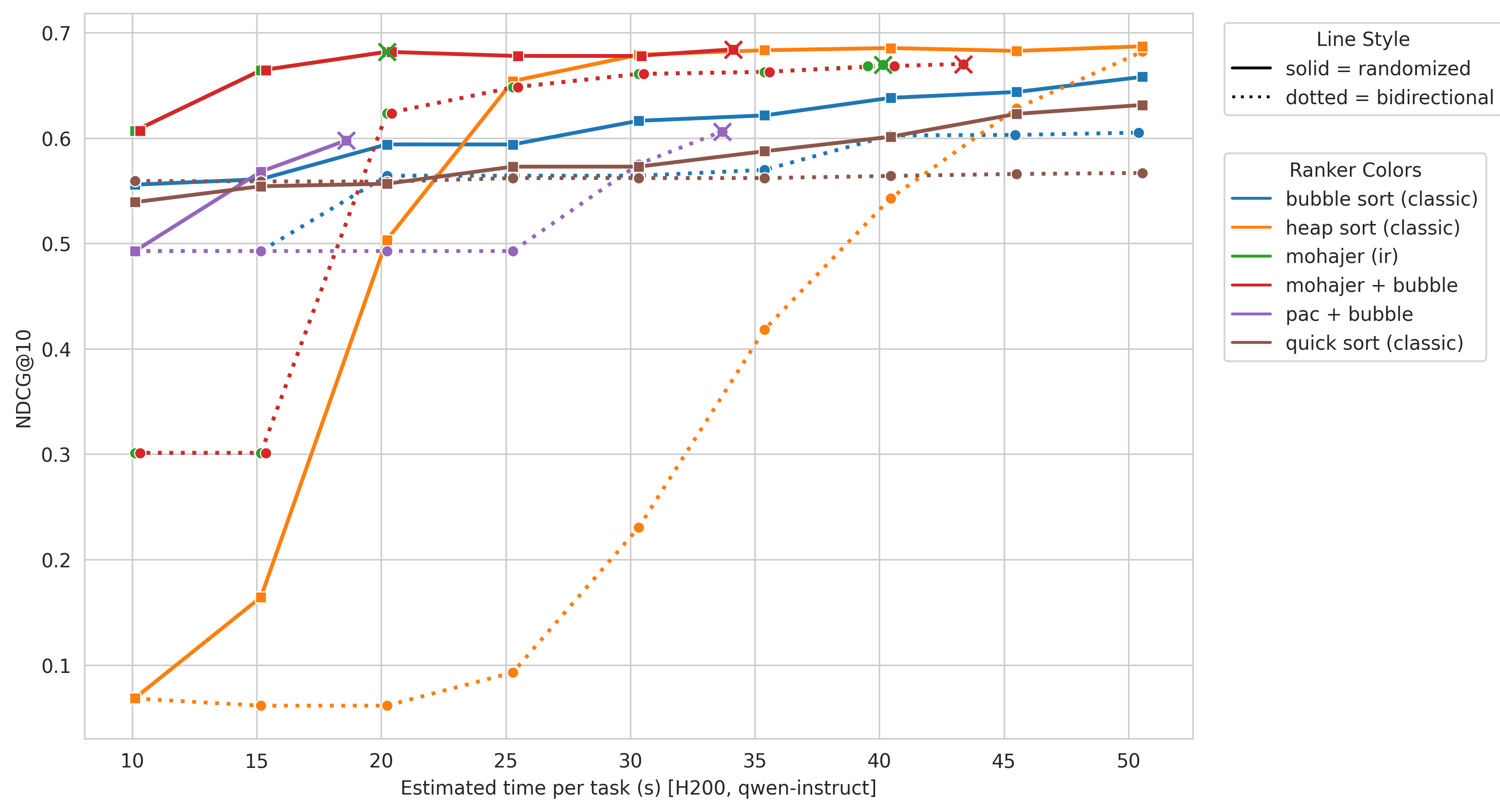}
    \caption{TREC DL 2019 and DL 2020 (Qwen3-4B-Instruct-2507): NDCG@10 vs estimated time per task across GPUs, with both oracles shown. Colors denote rankers; solid lines are randomized and dotted lines are bidirectional. X marks show when an algorithm has converged.}
    \label{fig:limit_time_all_qwen_all_gpus}
\end{figure}

\clearpage
\section{Statistical Significance}
\label{sec:significance}
\renewcommand{\thetable}{A.\arabic{table}}  

To quantify the stability and significance of the reported results, we conduct two complementary non-parametric bootstrap analyses. First, we quantify seed-resampling uncertainty: we evaluate each method under 8 different oracle seeds and report the 95\% confidence interval (CI) half-width of the mean NDCG@10 across seeds (10,000 resamples). These CIs are reported directly in Table~\ref{tab:budget_ndcg10_oracles_flan_xl} for the randomized-direction oracle (the bidirectional oracle is deterministic given pairwise outcomes). 

Second, we report paired bootstrap significance tests over queries. For each budget, we resample queries with replacement (10,000 resamples) and measure whether the mean difference in NDCG@10 between two methods is significantly different from zero ($p < 0.05$). Table~\ref{tab:sig_vs_bubble} reports tests comparing Mohajer+Bubble against BubbleSort; Table~\ref{tab:sig_vs_heapsort} compares Mohajer+Bubble against HeapSort.

\begin{table}[H]
\centering
\scriptsize
\begin{adjustbox}{max width=\linewidth}
\begin{tabular}{llrrrrrrrrr}
\toprule
& & \multicolumn{9}{c}{Number of LLM calls (budget)} \\
\cmidrule(lr){3-11}
Oracle & Comparison (A vs B) & 100 & 150 & 200 & 250 & 300 & 350 & 400 & 450 & 500 \\
\midrule
Bidirectional & Mohajer+Bubble vs BubbleSort     & $\downarrow$ (-19.2) & $\downarrow$ (-19.2) & $\uparrow$ (+5.9) & $\uparrow$ (+8.4) & $\uparrow$ (+9.7) & $\uparrow$ (+9.3) & $\uparrow$ (+6.6) & $\uparrow$ (+6.7) & $\uparrow$ (+6.5) \\
Randomized    & Mohajer+Bubble vs BubbleSort     & $\uparrow$ (+5.1) & $\uparrow$ (+10.3) & $\uparrow$ (+8.8) & $\uparrow$ (+8.4) & $\uparrow$ (+6.1) & $\uparrow$ (+6.2) & $\uparrow$ (+4.1) & $\uparrow$ (+3.5) & $\uparrow$ (+2.1) \\
\bottomrule
\end{tabular}
\end{adjustbox}
\caption{Paired bootstrap significance over queries for Mohajer+Bubble vs. BubbleSort on TREC DL19+DL20 (Flan-T5-XL). Each cell shows the direction and mean $\Delta$NDCG@10 difference. $\uparrow/\downarrow$ indicates a statistically significant difference (paired bootstrap over queries, 10,000 resamples, $p < 0.05$); $=$ indicates not significant.}
\label{tab:sig_vs_bubble}
\end{table}

\begin{table}[H]
\centering
\scriptsize
\begin{adjustbox}{max width=\linewidth}
\begin{tabular}{llrrrrrrrrr}
\toprule
& & \multicolumn{9}{c}{Number of LLM calls (budget)} \\
\cmidrule(lr){3-11}
Oracle & Comparison (A vs B) & 100 & 150 & 200 & 250 & 300 & 350 & 400 & 450 & 500 \\
\midrule
Bidirectional & Mohajer+Bubble vs HeapSort & $\uparrow$ (+23.3) & $\uparrow$ (+24.0) & $\uparrow$ (+56.2) & $\uparrow$ (+55.5) & $\uparrow$ (+43.0) & $\uparrow$ (+24.4) & $\uparrow$ (+12.5) & $\uparrow$ (+4.2) & $=$ (-1.2) \\
Randomized    & Mohajer+Bubble vs HeapSort & $\uparrow$ (+53.8) & $\uparrow$ (+50.0) & $\uparrow$ (+17.9) & $\uparrow$ (+2.4) & $=$ (-0.1) & $=$ (+0.1) & $=$ (-0.7) & $=$ (-0.4) & $=$ (-0.8) \\
\bottomrule
\end{tabular}
\end{adjustbox}
\caption{Paired bootstrap significance over queries for Mohajer+Bubble vs.\ HeapSort on TREC DL19+DL20 (Flan-T5-XL). Each cell shows the direction and mean $\Delta$NDCG@10 (A$-$B). $\uparrow/\downarrow$ indicates a statistically significant difference (paired bootstrap over queries, 10,000 resamples, $p < 0.05$); $=$ indicates not significant.}
\label{tab:sig_vs_heapsort}
\end{table}

\FloatBarrier
\clearpage
\onecolumn
\section{Proof of Aggregate Unbiasedness for Randomized-Direction Oracle}
\label{sec:unbiased-proof}
Here we show that, despite the individual order bias that is present in each inference call, the randomized-direction oracle still achieves comparison results that do not depend on input order. Namely, we show that the probability of preferring document A over document B does not depend on the order in which it is fed to the comparator, and thus fulfills the property $\Pr[V_{ij}=1] = 1 - \Pr[V_{ji}=1]$. Let $V_{ij}$ be the output of the randomized-direction oracle. By definition, the oracle selects the input order $(d_i, d_j)$ or $(d_j, d_i)$ with equal probability $p=0.5$. The probability of the oracle preferring $d_i$ over $d_j$ is given by:
\begin{align*}
    \Pr[V_{ij}=1] &= \frac{1}{2} \Pr[\text{LLM}(d_i, d_j) = 1] + \frac{1}{2} \Pr[\text{LLM}(d_j, d_i) = 0] \\
    &= \frac{1}{2} \left( 1 - \Pr[\text{LLM}(d_i, d_j) = 0] \right) + \frac{1}{2} \left( 1 - \Pr[\text{LLM}(d_j, d_i) = 1] \right) \\
    &= 1 - \frac{1}{2} \left( \Pr[\text{LLM}(d_j, d_i) = 1] + \Pr[\text{LLM}(d_i, d_j) = 0] \right) \\
    &= 1 - \Pr[V_{ji}=1]
\end{align*}
This result confirms that the oracle is reciprocal in expectation. While any single LLM inference may be biased toward a specific position, the randomization of the oracle ensures that the aggregate estimator is symmetric and unbiased with respect to the document order.

\FloatBarrier

\end{document}